\documentclass[runningheads]{llncs}

\usepackage[year=2026,ID=8612]{eccv}

\usepackage{eccvabbrv}

\usepackage{graphicx}
\usepackage{booktabs}

\usepackage[accsupp]{axessibility}  

\usepackage[utf8]{inputenc} 
\usepackage[T1]{fontenc}    
\usepackage{hyperref}       
\usepackage{url}            
\usepackage{amsfonts}       
\usepackage{nicefrac}       
\usepackage{microtype}      
\usepackage{xcolor}         

\usepackage{multirow}
\usepackage{diagbox}
\usepackage{amsmath}
\usepackage{subcaption}
\usepackage{makecell}

\usepackage{float}

\usepackage{hyperref}

\usepackage{orcidlink}

\begin{document}

\title{GD-FPS: Growth-Driven Feedforward Parameter Selection for Efficient Fine-Tuning}\vspace{-5pt}


\author{Kenneth Yang\inst{1,2} \and
Wen-Li Wei\inst{1} \and
Jen-Chun Lin\inst{1}}

\authorrunning{F.~Author et al.}

\institute{Academia Sinica, Taipei, TAIWAN \and
National Taiwan University, Taipei, TAIWAN\\
\email{\{s109062136,lilijinjin,jenchunlin\}@gmail.com}}
\vspace{-5pt}
\maketitle

\vspace{-10pt}\begin{abstract}


Parameter-Efficient Fine-Tuning (PEFT) has emerged as a key strategy for adapting large-scale pre-trained models to downstream tasks, but existing approaches face notable limitations. Addition-based methods, such as Adapters, introduce inference latency and engineering complexity, whereas selection-based methods like Gradient-based Parameter Selection (GPS) require a full backward pass. The reliance on gradients not only incurs massive memory usage and substantial computational latency, but also leaves the selection vulnerable to the randomness of stochastic batch sampling. To resolve this, we propose Growth-Driven Feedforward Parameter Selection (GD-FPS). Operating entirely via forward passes, this strictly gradient-free method identifies the optimal parameter subset by scaling intrinsic weight magnitudes by their relative activation growth against a pre-training anchor. Evaluated on $26$ visual tasks spanning image classification and semantic segmentation, GD-FPS achieves competitive or superior performance over state-of-the-art PEFT baselines. Crucially, compared to GPS, it reduces peak memory usage by nearly $18\times$ and accelerates execution by over $2.7\times$ during the parameter selection stage. By guaranteeing deterministic selection, GD-FPS offers a memory-efficient, fast, and robust solution for fine-tuning.\vspace{-4pt}
\vspace{-10pt}
\end{abstract}
\section{Introduction}\vspace{-3pt}
\label{sec:intro}




Large-scale pre-trained models have become a cornerstone of foundation model research. These models, developed by training on vast and varied datasets, demonstrate impressive general capabilities across a wide range of tasks. To unlock their full potential and achieve state-of-the-art results for each downstream task, a crucial technique is fine-tuning. This process adapts these general-purpose models to specific tasks or domains, enabling specialization such as fine-tuning BERT~\cite{devlin-etal-2019-bert} for sentiment analysis on movie reviews~\cite{he2023sensitivity} or adapting a Vision Transformer (ViT)~\cite{dosovitskiy2020image} to detect brain tumors from MRI scans~\cite{diagnostics13122094}, among many other downstream applications~\cite{tseng2024music, ruiz2023dreambooth, anisuzzaman2025fine}.



However, the naive approach of fine-tuning all model parameters—--commonly referred to as full fine-tuning—--is often impractical due to two primary challenges. First, it poses a highly complex optimization challenge, as billions of parameters must be updated using the relatively small datasets typical of downstream tasks, often resulting in suboptimal performance. Second, full fine-tuning is a computationally and resource-intensive endeavor due to its massive memory usage. The associated costs are substantial, including the storage of the model parameters themselves, their gradients during backpropagation, optimizer states, and crucially, the intermediate activations required for gradient computation. This renders full fine-tuning computationally and memory prohibitive, and ultimately impractical for many applications.


To mitigate the challenges of full fine-tuning, Parameter-Efficient Fine-Tuning (PEFT) has emerged as a leading approach. PEFT methods adapt models for downstream tasks by optimizing only a minimal set of parameters, often achieving performance that is comparable or even superior to full fine-tuning while incurring only a fraction of the computational cost.




Many prominent PEFT techniques are addition-based, meaning they introduce new, trainable components into the existing model. For example, Adapters~\cite{houlsby2019parameter} insert small, learnable neural modules between the model's layers. However, these methods generally face two significant issues. The first is the additional overhead they introduce, which can increase the model's storage size and inference latency. The second, and more critical, challenge is the engineering complexity, which arises from the considerable design difficulty of determining the optimal placement of these new modules~\cite{nowak2024optimaladapterplacementefficient, zhang2024gradient}. This decision often relies on task-specific heuristics; for instance, tasks requiring high-level semantic refinement might benefit from inserting Adapters in deeper layers, whereas tasks focused on low-level features may be better served by placing them in shallower layers. Such reliance on expert tuning and model-specific knowledge undermines their potential as a simple ``plug-and-play'' solution.

An alternative category, selection-based PEFT, has recently been proposed to address model overhead and engineering complexity. \emph{Our method falls into this group}, with Gradient-based Parameter Selection (GPS)~\cite{zhang2024gradient} being the closest prior work. The GPS process involves two stages: parameter selection followed by fine-tuning. To select parameters, it must first unfreeze the entire model to compute gradients for all parameters on the downstream task. Once the parameters with the highest gradient magnitudes are identified, only that subset is fine-tuned. This approach successfully eliminates additional parameters and the need for complex engineering design. 
However, the requirement to compute gradients for the full model during the selection stage contradicts the primary motivation of PEFT. It introduces substantial computational overhead and suffers from the same peak memory usage as full fine-tuning, which creates a dilemma. 
Furthermore, the reliance on batch-by-batch gradient calculations renders the selection process inherently noisy and highly sensitive to environment configurations, such as stochastic batch sampling and batch size limitations.

To simultaneously overcome these limitations in memory, latency, and robustness, we propose \emph{Growth-Driven Feedforward Parameter Selection (GD-FPS)}. By utilizing a strictly \emph{gradient-free approach}, GD-FPS completely bypasses the backward pass during selection, providing a genuinely accelerated, memory-efficient, and deterministic solution.

Concretely, the limitations of prior gradient-based methods stem directly from the mechanics of backpropagation: the backward pass itself drives the latency, storing full-model computation graphs dictates the memory bottleneck, and the reliance on sampled gradients injects stochastic noise. 

This raises a critical question: can we bypass this issue with a gradient-free approach? Why not determine parameter importance on-the-fly during the forward pass? This is the core idea behind our method. 
We propose quantifying parameter importance by scaling intrinsic weight magnitudes by their task-specific activation growth relative to a pre-training anchor. Relying purely on feedforward activations, this approach requires only forward passes, guaranteeing a fast, memory-efficient, and fully deterministic selection process.

We evaluate our approach across a comprehensive suite of 26 visual tasks, encompassing both image classification and semantic segmentation across different model architectures. Extensive experiments demonstrate that GD-FPS achieves comparable or even superior performance to the state-of-the-art GPS baseline, while crucially reducing peak GPU memory usage by $18\times$ and accelerating execution by over $2.7\times$ during the parameter selection stage. Coupled with a nearly deterministic selection process, GD-FPS establishes itself as a highly stable, resource-efficient, and strictly superior alternative.\vspace{-13pt}
\section{Related Work}\vspace{-8pt}
\label{sec:related_works}
Following prior work~\cite{zhang2024gradient}, we categorize PEFT methods into two main paradigms: \textbf{addition-based methods}, which augment the model with new trainable components, and \textbf{selection-based methods}, which identify and update a subset of the model’s original parameters.\vspace{-15pt}

\subsection{Addition-based Methods}\vspace{-3pt} 
Addition-based methods augment the model with new trainable components but typically incur two major drawbacks: increased inference overhead and substantial engineering complexity.

The first drawback, increased inference overhead, arises in multiple forms. Adapter-based methods and their variants~\cite{houlsby2019parameter, chen2022adaptformer, gao2024clip, gao2023llama, wang2020k} insert additional modules directly into the model architecture, thereby enlarging model size and resulting in slower inference. While Low-Rank Adaptation (LoRA)~\cite{hu2021lora} mitigates this latency by merging low-rank matrices back into the original weights, it fails to address the second—and more critical—design challenge.



The second drawback lies in the substantial engineering complexity, which limits these methods from being truly model- or task-agnostic. For approaches such as Adapters and LoRA, determining the optimal placement, dimensionality, and architecture of the new components presents a considerable design challenge that often depends on task-specific heuristics. Similarly, prompt-based methods, exemplified by Visual Prompt Tuning (VPT)~\cite{jia2022visual, ju2022prompting, hu2021knowledgeable, ding2021openprompt}, inject learnable context tokens into the input sequence, introducing their own design challenges while also incurring inference overhead by lengthening the input. This becomes problematic for models not designed to accommodate variable input lengths, thereby undermining their claim of universally applicable ``plug-and-play'' solutions.\vspace{-9pt}


\subsection{Selection-based Methods}\vspace{-2pt}
Selection-based methods, to which our proposed approach belongs, avoid introducing additional parameters. Early variants typically relied on simple heuristics, such as tuning only bias terms~\cite{zaken2021bitfit} or updating the final few layers of the network~\cite{houlsby2019parameter}. While straightforward, these methods often lag behind in performance. More recently, principled approaches have emerged to narrow this gap. The most relevant prior work is GPS~\cite{zhang2024gradient}, which first performs a forward–backward pass to identify parameter groups with the largest gradient magnitudes, and then fine-tunes them. This approach is effective, incurs no inference overhead, and circumvents the engineering design challenges inherent to addition-based methods. 
However, computing and storing full-model gradients during its selection process severely undermines the core motivation of PEFT. This reliance on the backward pass incurs the prohibitive peak memory usage of full fine-tuning, introduces substantial computational latency, and renders the selection process vulnerable to stochastic batch sampling. Our work directly resolves these three limitations. By proposing a purely gradient-free strategy that identifies parameters via forward passes, our method ensures a deterministic selection while maximizing both computational and memory efficiency.\vspace{-11pt}

\section{Preliminary of Selection-based Fine-tuning}\vspace{-4pt}


Selection-based fine-tuning updates only a sparse subset of parameters while freezing the rest. Empirical findings indicate that task adaptation operates in a substantially lower intrinsic dimensionality than the full parameter space~\cite{aghajanyan2020intrinsic}, motivating sparse parameter updates.

Formally, the objective of selection-based fine-tuning can be expressed as follows: Let $\theta_0 \in \mathbb{R}^P$ denote the pre-trained model parameters; the objective is to obtain fine-tuned parameters $\theta$ by solving:\vspace{-3pt}
\begin{equation}
\label{eq:constrained_optimization}
\min_{\theta} \mathcal{L}(\theta) \quad \text{subject to} \quad \|\theta - \theta_0\|_0 \le p.\vspace{-5pt}
\end{equation}
Here, $\|\cdot\|_0$ denotes the $\ell_0$-norm, which counts non-zero elements in the update vector $\Delta\theta = \theta - \theta_0$. The budget $p$ is a hyperparameter specifying the number of tunable parameters, with $p \ll P$. This constraint ensures only a small, selected subset is updated while keeping the architecture unchanged.

GPS~\cite{zhang2024gradient} identifies this sparse subset by assigning an importance score, $I_{\text{GPS}}$, to each weight, selecting the top-$k$ highest-scoring parameters within each neuron to maintain a balanced distribution across the network. For a given weight $w_{k,j}$, this score is defined as the sum of its gradient magnitudes over all mini-batches $\mathcal{B}$ from the downstream dataset $\mathcal{D}_{down}$:\vspace{-4pt}
\begin{equation}
\label{eq:gps_score}
    I_{\text{GPS}}(w_{k,j}) = \sum_{\mathcal{B} \in \mathcal{D}_{down}} \left\| \frac{\partial \mathcal{L}(\mathcal{B})}{\partial w_{k,j}} \right\|.
\end{equation}

\vspace{-4pt}Crucially, computing these scores requires a dedicated selection phase prior to any actual fine-tuning. During this phase, GPS must execute a complete epoch of both forward and backward passes over the dataset. This creates a severe memory bottleneck: because calculating these gradients necessitates unfreezing the entire model, the peak GPU memory usage becomes identical to that of full fine-tuning. Such heavy overhead fundamentally contradicts the core objective of PEFT, which is to provide a memory-efficient alternative for resource-constrained adaptation. Furthermore, this heavy reliance on gradient computation not only imposes substantial computational latency---making the selection phase prohibitively slow---but also renders the process highly sensitive to the inherent noise and variance of stochastic batch sampling.\vspace{-12pt}


\section{Method}\vspace{-5pt}
\label{sec:method}

\begin{figure}[t]
  \centering
  \includegraphics[width=\textwidth]{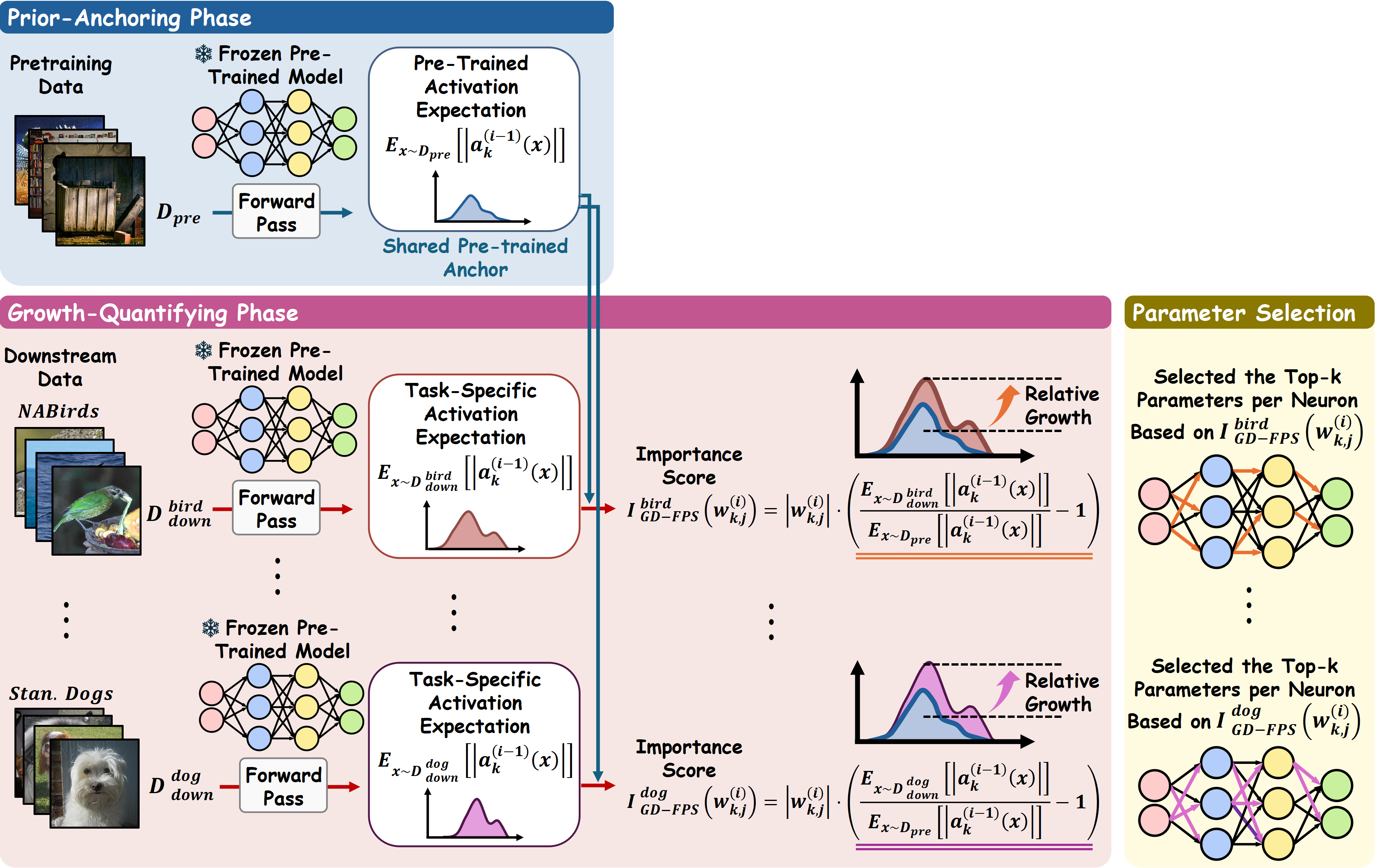}\vspace{-4pt}
  \caption{Overview of the proposed GD-FPS. GD-FPS evaluates parameter importance through a fast feedforward approach. By leveraging an activation baseline (anchor) that is computed only once using pre-training data (Prior-Anchoring Phase), it efficiently measures task-specific relative activation growth on the downstream dataset (Growth-Quantifying Phase). The final importance score is computed by scaling the intrinsic weight magnitude by this relative growth. Based on these scores, the top-$k$ parameters per neuron are selected for fine-tuning.}\vspace{-11pt}
  \label{fig:overview}
\end{figure}

\noindent\textbf{Framework overview.}
To resolve the dilemma presented by GPS, we design a framework that achieves the best of both paradigms: it strictly preserves the memory efficiency of standard PEFT while introducing a truly fast and robust parameter selection phase that identifies critical, task-specific parameters for fine-tuning. To this end, we propose GD-FPS, a purely gradient-free alternative. Rather than relying on computationally expensive backward passes, GD-FPS drives this selection process by quantifying activation growth during the forward pass between the pre-training and downstream domains.

As depicted in \cref{fig:overview}, GD-FPS operates in two phases. First, in the \emph{Prior-Anchoring Phase}, we process a subset of the pre-training data to establish a baseline (anchor) by computing the expected norm of the input activations corresponding to each weight. Second, before fine-tuning, the \emph{Growth-Quantifying Phase} evaluates the downstream dataset to measure the relative activation growth with respect to this pre-training anchor. GD-FPS then formulates a final importance score by scaling the intrinsic weight magnitude by this relative growth. We hypothesize that parameters exhibiting both high structural magnitude and significant task-specific activation growth contribute most to domain adaptation; accordingly, we select the top-k scoring parameters per neuron for fine-tuning. 

We first detail the formulation of our base Feedforward-based Parameter Selection (FPS) mechanism in \cref{sec:fps}. Building upon this foundation, we then introduce the full GD-FPS framework in \cref{sec:growth}.\vspace{-12pt}


\subsection{Feedforward-based Parameter Selection}\vspace{-2pt}
\label{sec:fps}

Using the loss gradient $\nabla_\theta \mathcal{L}(\theta)$ as a proxy for parameter importance requires constructing a full computational graph, leading to substantial computational overhead. To avoid this inefficiency, we instead derive importance scores directly from the forward pass. In this section, we formalize our base FPS mechanism, which forms the foundation of the full GD-FPS framework described in \cref{sec:growth}.

The fundamental operation of a neural network layer involves computing the output as the inner product of weights and input activations. Concretely, for a weight $w_{k,j}^{(i)}$ connecting the $k$-th neuron in layer $i-1$ to the $j$-th neuron in layer $i$, its direct contribution to the pre-activation output is given by $w_{k,j}^{(i)} \cdot a_k^{(i-1)}(x)$, where $a_k^{(i-1)}(x)$ denotes the input activation for a given sample $x$.

While the static weight magnitude $|w_{k,j}^{(i)}|$ is a well-established indicator of a parameter's intrinsic importance to the pre-trained representation~\cite{han2015learning, zhu2017prune, gale2019state, dettmers2023spqr}, it falls short when adapting to new domains. As Zhang \etal~\cite{zhang2024gradient} have empirically demonstrated, purely magnitude-based selection yields sub-optimal fine-tuning performance because it solely reflects pre-trained capacity while remaining blind to task-specific importance. In contrast, forward-pass activations provide dynamic, data-driven signals. Recent advances in foundation model compression further show that high-magnitude input activations serve as critical information hubs, significantly influencing a connection's structural contribution to the final output~\cite{yuan2022ptq4vit, dettmers2022llm, xiao2023smoothquant, li2023repq}.

Building on this, we formulate our base FPS importance score, denoted as $I_{FPS}(w_{k,j}^{(i)})$, which explicitly quantifies the joint magnitude of the static weight and its corresponding dynamic input activation over the downstream dataset. This evaluation is performed exclusively through forward passes on the downstream domain \textit{prior} to any fine-tuning:\vspace{-4pt}
\begin{equation}
\label{eq:fps_score}
I_{FPS}(w_{k,j}^{(i)}) = \mathbb{E}_{x \sim \mathcal{D}_{down}} \left[ |w_{k,j}^{(i)}| \cdot |a_k^{(i-1)}(x)| \right].
\end{equation}

\vspace{-4pt}This formulation enables efficient computation of parameter importance directly during the forward pass. By combining the model's pre-trained knowledge (via weight magnitude $|w_{k,j}^{(i)}|$) with data-specific properties from the downstream task (via activation magnitude $|a_k^{(i-1)}(x)|$), our method identifies critical parameters for adaptation. Consequently, this gradient-free design yields substantial efficiency gains without constructing a memory-intensive computational graph. Furthermore, because the expected activation magnitude is a global statistical property of the entire dataset, the final importance scores are theoretically invariant to batch size and stochastic sampling variance. We provide a detailed empirical validation of this robustness in \cref{sec:exp_classification}.\vspace{-14pt}


\subsection{The Growth-Driven Mechanism}\vspace{-4pt}
\label{sec:growth}

While the base FPS mechanism effectively identifies parameters that strongly influence the downstream outputs, its exclusive reliance on target-domain activations overlooks a fundamental premise of domain adaptation: the distribution shift between pre-training and downstream data. Mitigating the impact of these shifts is a central challenge in PEFT~\cite{kumar2022fine, ding2023parameter}, as true adaptation requires a model's internal activation patterns to dynamically change in response to new tasks.

This requirement exposes the core vulnerability of $I_{FPS}$, which calculates importance based solely on the magnitude of downstream activations.  Foundation models inherently learn general-purpose representations---such as generic geometric edge detectors or universal structural hubs---that yield massive activation signals regardless of the input domain~\cite{yosinski2014how, raghu2021do, dettmers2022llm}. $I_{FPS}$ evaluates the downstream data in a vacuum, making it biased toward these task-agnostic weights. If a parameter is predisposed to strong activations during pre-training, persistently high activations on the downstream task merely reflect its original, unchanged behavior, rather than actual adaptation to the new domain.

To identify the \textit{true} drivers of adaptation, we must look beyond these static pre-trained behaviors. We pinpoint the specific subset of parameters that are ``awakened'' during transfer---the critical weights whose activation profiles exhibit a profound growth in response to the new data distribution. To explicitly quantify this task-specific adaptation, we introduce our \textit{Growth-Driven (GD)} mechanism. Rather than evaluating downstream activations in isolation, we measure their relative growth against the model's original pre-training anchor.

As illustrated in \cref{fig:overview}, our GD-FPS framework operates in two distinct phases. The first is the \textit{Prior-Anchoring Phase}. Similar to the base FPS method, we perform a forward pass to compute activation magnitudes, but we execute this on a randomly sampled subset of the original pre-training data, denoted as $\mathcal{D}_{pre}$. The sole objective of this phase is to establish a behavioral baseline by applying the activation expectation formulation from \cref{eq:fps_score} to the pre-training domain. Specifically, for any weight $w_{k,j}^{(i)} \in \theta_0$, we compute its pre-trained activation expectation as $\mathbb{E}_{x \sim \mathcal{D}_{pre}} [|a_k^{(i-1)}(x)|]$. Crucially, this phase is a one-time procedure used to construct a universal baseline. Even if the model is subsequently fine-tuned on various downstream tasks, this single pre-trained anchor is shared across all of them. Our implementation requires merely $12,800$ images for $\mathcal{D}_{pre}$, rendering the amortized computational overhead entirely negligible. 

The second phase is the \textit{Growth-Quantifying Phase}. Here, we evaluate the model on the downstream target dataset, $\mathcal{D}_{down}$, to obtain the corresponding task-specific activation expectation, $\mathbb{E}_{x \sim \mathcal{D}_{down}} [|a_k^{(i-1)}(x)|]$. Finally, rather than relying solely on the absolute downstream signal, our proposed importance score $I_{GD-FPS}$ evaluates each weight by calculating its relative growth from the pre-training anchor:\vspace{-9pt}
\begin{equation}
\label{eq:gd_fps}
I_{GD-FPS}(w_{k,j}^{(i)}) = |w_{k,j}^{(i)}| \cdot \left( \frac{\mathbb{E}_{x \sim \mathcal{D}_{down}} \left[ |a_k^{(i-1)}(x)| \right]}{\mathbb{E}_{x \sim \mathcal{D}_{pre}} \left[ |a_k^{(i-1)}(x)| \right]} - 1 \right).
\end{equation}

\vspace{-5pt}By formulating the score as a relative growth, $I_{GD-FPS}$ captures the specific growth of the downstream signal relative to its pre-training anchor. This ensures that our framework ignores static, task-agnostic importance and instead prioritizes weights that exhibit a significant task-specific activation surge.\vspace{-11pt}





\section{Experiments}\vspace{-3pt}
\label{sec:experiments}
We evaluate the proposed GD-FPS across diverse downstream tasks and model backbones, encompassing both image classification (\cref{sec:exp_classification}) and semantic segmentation (\cref{sec:exp_segmentation}). Finally, we systematically study the impact of our core design choices through comprehensive ablation studies (\cref{sec:ablation}).\vspace{-11pt}

\subsection{Image Classification}\vspace{-2pt}
\label{sec:exp_classification}

\subsubsection{Experimental settings.} Following prior work~\cite{zhang2024gradient, jia2022visual, lian2022scaling}, we utilize a Vision Transformer (ViT-B/16)~\cite{dosovitskiy2020image} pre-trained on ImageNet-21K~\cite{deng2009imagenet} to evaluate the proposed GD-FPS on $24$ image classification tasks across two benchmarks: (i) the Fine-Grained Visual Classification (FGVC) benchmark, which includes $5$ datasets (CUB-200-2011~\cite{wah2011caltech}, NABirds~\cite{van2015building}, Oxford Flowers~\cite{nilsback2008automated}, Stanford Dogs~\cite{khosla2011novel}, and Stanford Cars~\cite{gebru2017fine}); and (ii) the VTAB-1k benchmark~\cite{zhai2019large}, which consists of $19$ diverse tasks categorized into \textit{Natural}, \textit{Specialized}, and \textit{Structured} groups, each containing exactly $1,000$ training examples.\vspace{-12pt}

\sloppy
\subsubsection{Implementation details.} To ensure a strictly fair comparison, all configurations---including data splits, optimizer, learning rate scheduler, implementation environment, and the number of tunable parameters---are kept identical to our primary baseline, GPS~\cite{zhang2024gradient}. All experiments are conducted on NVIDIA RTX A6000 GPUs. For our GD-FPS framework, the \emph{Prior-Anchoring Phase} randomly samples $12,800$ images directly from the pre-training dataset to capture the baseline activation expectation. Executing this phase requires merely $27.38$ seconds on a single GPU. As a one-time procedure shared across all downstream datasets, the amortized computational overhead of this phase is entirely negligible.\vspace{-12pt}

\subsubsection{Baseline methods.} We compare GD-FPS against established fine-tuning paradigms, which can be categorized into: (i) \textbf{Full fine-tuning}, which updates all model parameters but incurs prohibitive computational costs; (ii) \textbf{Selection-based methods}, which fine-tune a sparse subset of existing weights, including standard approaches like Linear probing~\cite{jia2022visual} and Bias tuning~\cite{zaken2021bitfit}, as well as GPS~\cite{zhang2024gradient}, which serves as our primary state-of-the-art competitor; and (iii) \textbf{Addition-based methods}, which introduce new trainable modules that either add permanent inference overhead (Adapter~\cite{houlsby2019parameter}, VPT~\cite{jia2022visual}, SPT-Adapter~\cite{he2023sensitivity}) or can be reparameterized into the backbone during inference (LoRA~\cite{hu2021lora}, SSF~\cite{lian2022scaling}, SPT-LoRA~\cite{he2023sensitivity}).

\subsubsection{Performance on image classification.} \cref{tab:fgvc,tab:vtab} summarize the results across both benchmarks. On FGVC, our GD-FPS achieves state-of-the-art performance with a mean accuracy of $92.07\%$, outperforming the closest competing methods, GPS ($+0.29\%$) and SSF ($+1.35\%$). On the highly diverse VTAB-1k benchmark, GD-FPS attains a competitive mean accuracy of $75.04\%$. While performing on par with GPS ($75.18\%$), it maintains a substantial margin over the next best approach, SPT-LoRA ($+0.97\%$). Crucially, while GD-FPS delivers comparable or superior classification accuracy relative to GPS, it simultaneously provides significant advantages in computational efficiency and robustness during the parameter selection phase, as detailed in the following analysis.\vspace{-8pt}

  \begin{table}[t]  
    \centering
    \caption{Performance comparison on FGVC. Bold best, underline second.}\vspace{-6pt}
\setlength{\tabcolsep}{4pt} 
\scalebox{0.66}{ 
    \begin{tabular}{l|ccccc|cc}
    \toprule
    \diagbox{Method}{Dataset}&\makecell[c]{CUB\\-2011} & \makecell[c]{NA-\\Birds} & \makecell[c]{Oxford\\Flowers} & \makecell[c]{Stan.\\Dogs} & \makecell[c]{Stan.\\Cars} & \makecell[c]{Mean\\Acc.} & \makecell[c]{Params.\\(\%)} \\
    \midrule
    Full Fine-Tuning~\cite{jia2022visual}        & 87.3          & 82.7    & 98.8              & 89.4          & 84.5          & 88.54     & 100.00 \\
    \midrule
    Linear~\cite{jia2022visual}      & 85.3          & 75.9    & 97.9              & 86.2          & 51.3          & 79.32     & 0.21   \\
    Bias~\cite{zaken2021bitfit}          & 88.4          & 84.2    & 98.8              & 91.2          & 79.4          & 88.40     & 0.33   \\
    \midrule
    Adapter~\cite{houlsby2019parameter}    & 87.1          & 84.3    & 98.5              & 89.8          & 68.6          & 85.66     & 0.48   \\
    LoRA~\cite{hu2021lora}         & 85.6          & 79.8    & 98.9              & 87.6          & 72.0          & 84.78     & 0.90   \\
    VPT-Shallow~\cite{jia2022visual}   & 86.7          & 78.8    & 98.4              & 90.7          & 68.7          & 84.62     & 0.29   \\
    VPT-Deep~\cite{jia2022visual}      & 88.5          & 84.2    & 99.0              & 90.2          & 83.6          & 89.11     & 0.99   \\
    SSF~\cite{lian2022scaling}           & 89.5        & 85.7    & \underline{99.6}  & 89.6          & 89.2          & 90.72     & 0.45   \\
    SPT-Adapter~\cite{he2023sensitivity}   & 89.1          & 83.3    & 99.2              & 91.1          & 86.2          & 89.78     & 0.47   \\
    SPT-LoRA~\cite{he2023sensitivity}      & 88.6          & 83.4    & 99.5              & 91.4  & 87.3          & 90.04     & 0.60   \\
    \midrule
    GPS~\cite{zhang2024gradient}   & \textbf{89.9} & \underline{86.7} & \textbf{99.7}     & \textbf{92.2} & \underline{90.4}  & \underline{91.78} & 0.77 \\
    GD-FPS (Ours)    & \underline{89.7} & \textbf{86.8} & \textbf{99.7}     & \underline{92.0} & \textbf{92.2}  & \textbf{92.07} & 0.77 \\
    \bottomrule
    \end{tabular}
}
    \vspace{2pt} 
    \label{tab:fgvc}
  \end{table}

\begin{table}[t]
    \centering
    \caption{Performance comparison on VTAB-1k.}\vspace{-5pt}
    
\resizebox{\columnwidth}{!}{
    \begin{tabular}{l|ccccccc|cccc|cccccccc|cc}
    \toprule
    \multirow{2}{*}{\diagbox[height=7\line]{Method \\ \\ \\}{\\ \\ \\ Dataset}} & \multicolumn{7}{c}{Natural} & \multicolumn{4}{c}{Specialized} & \multicolumn{8}{c}{Structured} & \multicolumn{2}{c}{VTAB}\\
    \cline{2-22}
    & \rotatebox{90}{CIFAR-100} & \rotatebox{90}{Caltech101} & \rotatebox{90}{DTD} & \rotatebox{90}{Flowers102} & \rotatebox{90}{Pets} & \rotatebox{90}{SVHN} & \rotatebox{90}{Sun397} & \rotatebox{90}{Patch Camelyon} & \rotatebox{90}{EuroSAT} & \rotatebox{90}{Resisc45} & \rotatebox{90}{Retinopathy} & \rotatebox{90}{Clevr/count} & \rotatebox{90}{Clevr/distance} & \rotatebox{90}{DMLab} & \rotatebox{90}{KITTI/distance} & \rotatebox{90}{dSprites/loc} & \rotatebox{90}{dSprites/ori} & \rotatebox{90}{SmallNORB/azi} & \rotatebox{90}{SmallNORB/ele} & \rotatebox{90}{Mean Acc.} & \rotatebox{90}{Mean Params. (\%)} \\
    \midrule 
    Full Fine-Tuning~\cite{jia2022visual} & 68.9 & 87.7 & 64.3 & 97.2 & 86.9 & 87.4 & 38.8 & 79.7 & 95.7 & 84.2 & 73.9 & 56.3 & 58.6 & 41.7 & 65.5 & 57.5 & 46.7 & 25.7 & 29.1 & 65.57 & 100.00 \\
    \midrule
    Linear~\cite{jia2022visual} & 63.4 & 85.0 & 64.3 & 97.0 & 86.3 & 36.6 & 51.0 & 78.5 & 87.5 & 68.6 & 74.0 & 34.3 & 30.6 & 33.2 & 55.4 & 12.5 & 20.0 & 9.6 & 19.2 & 53.00 & 0.05 \\
    Bias~\cite{zaken2021bitfit} & 72.8 & 87.0 & 59.2 & 97.5 & 85.3 & 59.9 & 51.4 & 78.7 & 91.6 & 72.9 & 69.8 & 61.5 & 55.6 & 32.4 & 55.9 & 66.6 & 40.0 & 15.7 & 25.1 & 62.05 & 0.16 \\
    \midrule
    Adapter~\cite{houlsby2019parameter} & 74.1 & 86.1 & 63.2 & 97.7 & 87.0 & 34.6 & 50.8 & 76.3 & 88.0 & 73.1 & 70.5 & 45.7 & 37.4 & 31.2 & 53.2 & 30.3 & 25.4 & 13.8 & 22.1 & 55.82 & 0.31 \\
    LoRA~\cite{hu2021lora} & 68.1 & 91.4 & 69.8 & 99.0 & 90.5 & 86.4 & 53.1 & 85.1 & 95.8 & 84.7 & 74.2 & \underline{83.0} & 66.9 & 50.4 & 81.4 & 80.2 & 46.6 & 32.2 & 41.1 & 72.63 & 0.90 \\
    VPT-Shallow~\cite{jia2022visual} & 77.7 & 86.9 & 62.6 & 97.5 & 87.3 & 74.5 & 51.2 & 78.2 & 92.0 & 75.6 & 72.9 & 50.5 & 58.6 & 40.5 & 67.1 & 68.7 & 36.1 & 20.2 & 34.1 & 64.85 & 0.13 \\
    VPT-Deep~\cite{jia2022visual} & {78.8} & 90.8 & 65.8 & 98.0 & 88.3 & 78.1 & 49.6 & 81.8 & \underline{96.1} & 83.4 & 68.4 & 68.5 & 60.0 & 46.5 & 72.8 & 73.6 & 47.9 & \underline{32.9} & 37.8 & 69.43 & 0.70 \\
    SSF~\cite{lian2022scaling} & 69.0 & 92.6 & {75.1} & \textbf{99.4} & \textbf{91.8} & {90.2} & 52.9 & \underline{87.4} & 95.9 & \underline{87.4} & 75.5 & 75.9 & 62.3 & 53.3 & 80.6 & 77.3 & \underline{54.9} & 29.5 & 37.9 & 73.10 & 0.28 \\
    SPT-Adapter~\cite{he2023sensitivity} & 72.9 & 93.2 & 72.5 & \underline{99.3} & 91.4 & 88.8 & \textbf{55.8} & 86.2 & \underline{96.1} & 85.5 & 75.5 & \underline{83.0} & \textbf{68.0} & 51.9 & 81.2 & 51.9 & 31.7 & \textbf{41.2} & \textbf{61.4} & 73.03 & 0.44 \\
    SPT-LoRA~\cite{he2023sensitivity} & 73.5 & {93.3} & 72.5 & \underline{99.3} & 91.5 & 87.9 & \underline{55.5} & 85.7 & \textbf{96.2} & 85.9 & \underline{75.9} & \textbf{84.4} & \underline{67.6} & 52.5 & 82.0 & 81.0 & 51.1 & 30.2 & 41.3 & {74.07} & 0.63 \\
    \midrule
    GPS~\cite{zhang2024gradient} & \textbf{81.1} & \textbf{94.2} & \textbf{75.8} & \textbf{99.4} & \underline{91.7} & \textbf{91.6} & 52.4 & \textbf{87.9} & \textbf{96.2} & {86.5} & \textbf{76.5} & 79.9 & 62.6 & \textbf{55.0} & \textbf{82.4} & \underline{84.0} & \textbf{55.4} & 29.7 & \underline{46.1} & \textbf{75.18} & 0.25 \\
    GD-FPS (Ours) & \underline{80.9} & \underline{93.5} & \underline{75.7} & \textbf{99.4} & 91.6 & \underline{91.5} & 51.1 & {87.1} & {95.5} & \textbf{88.1} & \underline{75.9} & 78.8 & 63.2 & \underline{54.5} & \underline{82.3} & \textbf{85.9} & {54.2} & 31.3 & 45.0 & \underline{75.04} & 0.25 \\
    \bottomrule
    \end{tabular}
}

    \vspace{-5pt} 
    \label{tab:vtab}
\end{table}

\subsubsection{Efficiency analysis.} 
\begin{figure}[t]
    \centering
    \includegraphics[width=\linewidth]{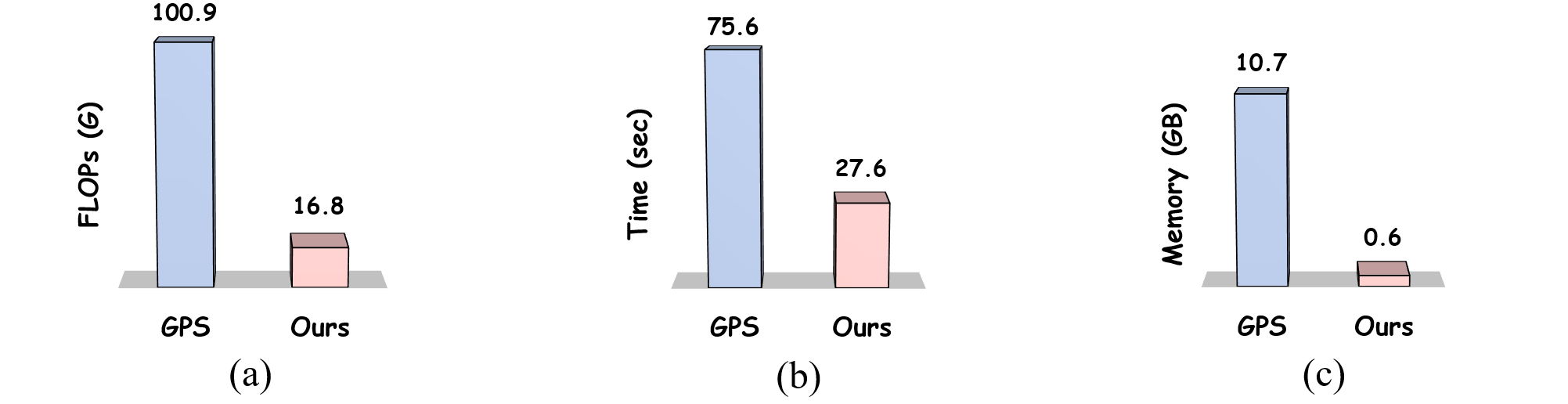}\vspace{-11pt} 
    \caption{Computational overhead on FGVC during the parameter selection phase: (a) FLOPs per image, (b) average selection latency, and (c) peak GPU memory usage.}\vspace{-12pt}
    \label{fig:ECCV_performance_combined}
\end{figure}

As shown in \cref{fig:ECCV_performance_combined}, a major advantage of GD-FPS is its substantially reduced computational overhead compared to GPS during the selection phase. Specifically, our method reduces peak GPU memory usage by nearly $18\times$, accelerates parameter selection by over $2.7\times$, and decreases computational cost from $100.9$ GFLOPs to $16.8$ GFLOPs per image. These significant efficiency gains stem directly from our gradient-free selection strategy, which entirely circumvents the heavy gradient-based iterations required by GPS.

\subsubsection{Robustness analysis.} Complementing the significant efficiency gains discussed above, a critical criterion for a reliable selective fine-tuning method is selection robustness: for a given model and dataset, the algorithm should reliably identify an identical subset of parameters for tuning, regardless of stochastic training factors. To demonstrate the superior robustness of GD-FPS, we first examine its theoretical stability. Recall the GD-FPS importance score defined in \cref{eq:gd_fps}. The core component of this score is the expected activation magnitude, which is a global statistical property of the dataset. For a dataset $\mathcal{D}$ partitioned into batches $\mathcal{B}$, this expectation can be exactly computed as a running average:\vspace{-4pt}
\begin{equation}
\label{eq:expected_activation}
\mathbb{E}_{x \sim \mathcal{D}} \left[ |a_k^{(i-1)}(x)| \right] = \frac{1}{|\mathcal{D}|} \sum_{\mathcal{B} \in \mathcal{D}} \sum_{x \in \mathcal{B}} |a_k^{(i-1)}(x)|\vspace{-3pt}
\end{equation}
Because this operation is a simple summation and averaging process over the entire dataset, the final computed importance score is strictly deterministic, remaining theoretically invariant to stochastic training factors.

In contrast, the core component of the baseline GPS~\cite{zhang2024gradient} importance score, $I_{\text{GPS}}$, is the gradient norm evaluated on individual batches, $\left\| \frac{\partial \mathcal{L}(\mathcal{B})}{\partial w_{k,j}} \right\|$, as defined in \cref{eq:gps_score}. This heavy reliance on batch-level gradients renders the GPS importance score highly vulnerable to local minima, the specific partitioning of batches, and the inherent stochasticity of the dataloader.

To empirically validate our theoretical claims, we introduce the \textbf{Intersection Ratio (IR)} to quantify selection robustness. Specifically, the IR measures the fraction of parameters that are consistently selected across different runs. As shown in \cref{fig:robustness_analysis}, we test robustness on the CUB-2011 and NABirds datasets under three conditions, defining the specific IR formulation for each given a fixed parameter budget $K$:\vspace{-3pt}
\begin{figure}[t]
    \centering
    \includegraphics[width=\linewidth]{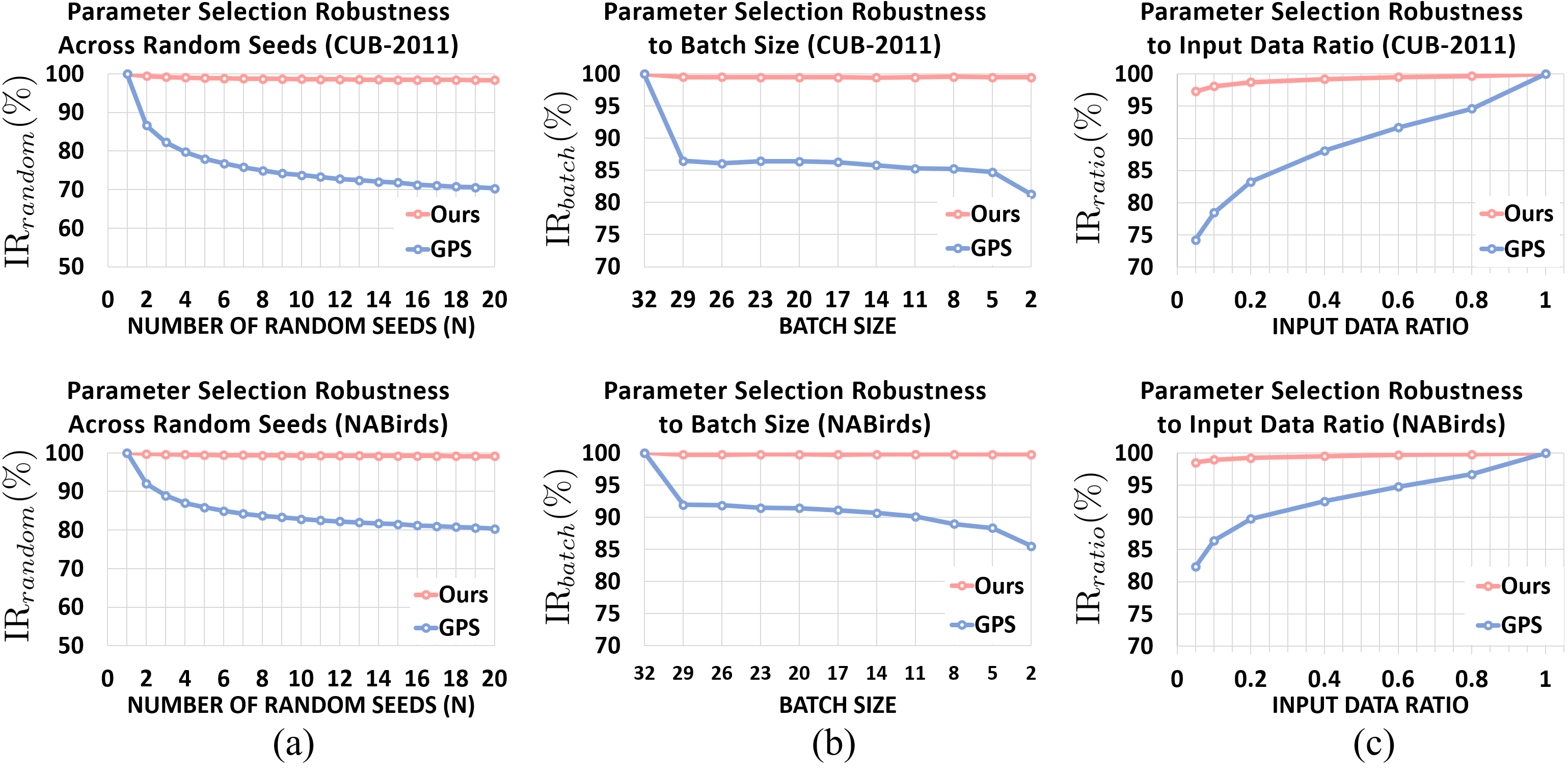}\vspace{-12pt}
    \caption{Robustness test of parameter selection between GD-FPS and GPS on the CUB-2011 and NABirds datasets. We report the intersection ratio of selected weights under (a) different numbers of random seeds, (b) different batch sizes, and (c) different available input data ratios for selection.}\vspace{-11pt}
    \label{fig:robustness_analysis}
\end{figure}
\begin{itemize}
    \item \textbf{Random seeds:} We run both methods $N$ times ($N \in [1, 20]$) using varying random seeds for data sampling. We measure the strict robustness across all trials by calculating the intersection of all $N$ selected parameter subsets $S_1, S_2, \dots, S_N$:\vspace{-3pt}
    \begin{equation}
        \text{IR}_{random} = \frac{|\bigcap_{n=1}^N S_n|}{K}.\vspace{-1pt}
    \end{equation}    
    \item \textbf{Batch size variations:} We define the parameter selection derived from a full-batch execution as the golden standard, denoted as $S_{gold}$. For a parameter subset $S_{batch}$ selected using progressively smaller batch sizes, the robustness is measured directly against this standard:\vspace{-4pt}
    \begin{equation}
        \text{IR}_{batch} = \frac{|S_{batch} \cap S_{gold}|}{K}.\vspace{-4pt}
    \end{equation}    
    \item \textbf{Input data ratios:} We randomly draw a reduced fraction of the dataset, $\mathcal{D}_{down}' \subset \mathcal{D}_{down}$, across various drop ratios to test stability under data scarcity. Let $S_{ratio}$ and $S_{gold}$ denote the parameter sets selected based on $\mathcal{D}_{down}'$ and the full dataset $\mathcal{D}_{down}$, respectively. The robustness of $S_{ratio}$ is measured directly against the full-data standard, $S_{gold}$, via the IR:\vspace{-4pt}
        \begin{equation}
            \text{IR}_{ratio} = \frac{|S_{ratio} \cap S_{gold}|}{K}.\vspace{-1pt}
        \end{equation}        
\end{itemize}

\vspace{-2pt}As illustrated in \cref{fig:robustness_analysis}, across all evaluated scenarios on both the CUB-2011 and NABirds datasets, GD-FPS consistently maintains an IR exceeding $97\%$. Remarkably, this near-perfect robustness holds even under extreme conditions, such as reducing the available input data ratio to a mere $5\%$, executing across $20$ distinct random seeds, or minimizing the batch size to $2$. In stark contrast, the IR of GPS drops by $20\% \sim 30\%$ under these same variations, highlighting the inherent brittleness of gradient-based selection methods.\vspace{-12pt}

\subsection{Semantic Segmentation}\vspace{-4pt}
\label{sec:exp_segmentation}
Semantic segmentation remains a notoriously challenging task for PEFT methods. Unlike image-level classification, segmentation requires models to develop fine-grained, pixel-level perceptual capabilities. This inherent shift in perceptual granularity complicates the fine-tuning process, historically making full fine-tuning a formidable and hard-to-beat baseline. To evaluate the universality and generalizability of our proposed GD-FPS, we extend our experiments to semantic segmentation benchmarks using a Swin-L backbone~\cite{liu2021swin}.

\subsubsection{Experimental setup.} Following prior work~\cite{zhou2025progressive}, we adopt a Swin-L~\cite{liu2021swin} pre-trained on ImageNet-21K~\cite{deng2009imagenet} as the backbone and UperNet~\cite{xiao2018unified} as the segmentation head. For a comprehensive comparison with state-of-the-art methods, we report the mean Intersection over Union (mIoU) and mean Accuracy (mAcc) on the validation sets, where higher values indicate better performance.\vspace{-12pt} 

\subsubsection{Datasets.} Consistent with existing literature~\cite{zhou2025progressive}, we comprehensively evaluate GD-FPS on two standard segmentation benchmarks: PASCAL VOC2012~\cite{everingham2015pascal} and Cityscapes~\cite{cordts2016cityscapes}. PASCAL VOC2012 consists of $1,464$ training and $1,449$ validation images spanning $21$ categories, serving as a key benchmark for limited-scale data scenarios. Cityscapes contains finely annotated images across $19$ categories, with $2,975$ training and $500$ validation images, and is widely adopted for evaluating urban scene understanding.\vspace{-12pt}

\subsubsection{Baselines.} We compare our method against baseline results reported in prior work~\cite{zhou2025progressive}, which encompass widely adopted PEFT techniques such as Bias tuning~\cite{zaken2021bitfit}, VPT~\cite{jia2022visual}, LoRA~\cite{hu2021lora}, and AdaptFormer~\cite{chen2022adaptformer}. We also compare against the recent state-of-the-art method, ProPETL~\cite{zhou2025progressive}. Belonging to the family of addition-based methods, ProPETL is specifically designed for dense prediction by modeling semantic segmentation as a granularity transfer process. It employs a progressive two-stage training paradigm: it first trains an adapter-like module on a midstream classification task, then freezes this module and concatenates a newly initialized structure trained for the downstream segmentation task.\vspace{-12pt}

\subsubsection{Performance on semantic segmentation.}
\begin{table}[t]
\centering
\caption{Performance comparison of different PEFT methods on semantic segmentation benchmarks. The symbol $^*$ denotes the learnable parameters in the backbone. The ``Freeze'' setting indicates that only the segmentation head is trained.}\vspace{-7pt}
\label{tab:segmentation_results}
\resizebox{.64\textwidth}{!}{%
\begin{tabular}{l ccccccc} 
\toprule
& \multicolumn{2}{c}{VOC2012} & \multicolumn{2}{c}{Cityscapes} & \multicolumn{2}{c}{Mean} & Param.$^*$ \\
\cmidrule(lr){2-3} \cmidrule(lr){4-5} \cmidrule(lr){6-7}
Method & mIoU & mAcc & mIoU & mAcc & mIoU & mAcc & (M) \\
\midrule
{Full fine-tuning\cite{zhou2025progressive}} & {84.38} & {89.82} & \underline{82.36} & \underline{88.53} & {83.37} & {89.18} & {195.00} \\
Freeze\cite{zhou2025progressive}           & 83.32 & 89.16 & 75.54 & 82.91 & 79.43 & 86.04 & 0      \\
Bias\cite{zaken2021bitfit}           & 84.25 & 90.25 & 77.73 & 84.89 & 80.99 & 87.57 & 0.30   \\
VPT \cite{jia2022visual}             & 85.69 & 91.12 & 78.95 & 86.56 & 82.32 & 88.84 & 3.61   \\
AdaptFormer\cite{chen2022adaptformer}     & 85.41 & 90.93 & 79.41 & 86.03 & 82.41 & 88.48 & 2.64   \\
LoRA\cite{hu2021lora}             & 85.36 & 91.34 & 80.54 & 87.23 & 82.95 & 89.29 & 4.55   \\
ProPETL\cite{zhou2025progressive}          & \textbf{86.11} & \textbf{92.08} & \textbf{81.67} & \textbf{88.23} & \textbf{83.89} & \textbf{90.16} & 3.30   \\
GP-FPS (Ours) & \underline{86.00} & \underline{91.65} & {81.24} & {87.91} & \underline{83.62} & \underline{89.78} & 2.60   \\
\bottomrule
\end{tabular}
}\vspace{-6pt}
\end{table}
We present the quantitative results for both VOC2012 and Cityscapes in \cref{tab:segmentation_results}. As demonstrated, GD-FPS achieves performance closely trailing the task-specific state-of-the-art, ProPETL, while establishing a substantial margin over other general-purpose PEFT methods. Crucially, among these general PEFT approaches, our GD-FPS is the only one to successfully surpass the formidable \textit{Full fine-tuning} baseline in both Mean mIoU and Mean mAcc.

By eliminating the need for a complex midstream transfer stage, our method avoids extensive training overhead. GD-FPS is trained for an average of $60$k iterations, whereas ProPETL requires approximately $90$k iterations, resulting in about $1.5\times$ increase in training iterations. It is also worth noting that while ProPETL is heavily engineered specifically for semantic segmentation, its granularity transfer concept is orthogonal to our approach and could easily be integrated into future extensions of GD-FPS. Overall, our results on semantic segmentation demonstrate that GD-FPS successfully transfers across different backbones and task domains, consistently achieving superior performance over existing PEFT methods.\vspace{-10pt}

\subsection{Ablation Studies}\vspace{-2pt}
\label{sec:ablation}

In this section, we present ablation studies on our proposed GD-FPS. We first examine the core design choices of the base FPS, including parameter selection schemes, norms for computing activation magnitude, and joint weight–activation consideration. Subsequently, we demonstrate the performance improvements introduced by the GD mechanism, specifically highlighting the necessity of our relative-growth formulation. Results are summarized in \cref{tab:ablation_vtab}.\vspace{-12pt}

\subsubsection{Parameter selection scheme and norm choice.}
Following GPS\cite{zhang2024gradient}, we examine FPS under different parameter selection schemes: layer-level and neuron-level. Specifically, layer-level selection identifies the top parameters with the highest importance scores within each layer, whereas neuron-level selection enforces this constraint for each individual neuron. In short, the layer-level scheme ensures at least one parameter is selected per layer, while the neuron-level scheme guarantees parameter selection across all neurons; however, under layer-level selection, some neurons may end up with no selected parameters.

\begin{table}[t]
    \centering
    \caption{Ablation study of FPS and GD-FPS on VTAB-1k. (a) layer-level selection on FPS. (b) neuron-level selection on FPS. (c) neuron-level selection without weight magnitude consideration on FPS. (d) Our full GD-FPS method with or without normalization. The number of tunable parameters is kept fixed across all experiments.}\vspace{-7pt}
    \label{tab:ablation_vtab}
    \setlength{\tabcolsep}{6pt} 
    \scalebox{0.73}{ 
        \begin{tabular}{c|l|c}
        \toprule
        \multicolumn{2}{l|}{\textbf{Method}} & \makecell[c]{\textbf{VTAB-1k} \\ \textbf{Mean Acc. (\%)}} \\
        \midrule
        \multirow{2}{*}{(a)} & FPS (layer-level + $\ell_1$-based) & 74.74 \\
                             & FPS (layer-level + $\ell_2$-based) & 74.64 \\
        \midrule
        \multirow{2}{*}{(b)} & FPS (neuron-level + $\ell_1$-based) & \underline{74.76} \\
                             & FPS (neuron-level + $\ell_2$-based) & 74.73 \\
        \midrule
        \multirow{2}{*}{(c)} & FPS (neuron-level + $\ell_1$-based w/o weight magnitude) & 74.43 \\
                             & FPS (neuron-level + $\ell_2$-based w/o weight magnitude) & 74.37 \\
        \midrule
        \multirow{2}{*}{(d)} & GD-FPS (w/o normalization) & 74.61 \\
                             & GD-FPS (ours) & \textbf{75.04} \\
        \bottomrule
        \end{tabular}
    }\vspace{-6pt}
\end{table}

For calculating the expected activation magnitude, we evaluate both an $\ell_1$-based and an $\ell_2$-based importance score. The $\ell_1$ formulation is the primary metric adopted in our approach (see \cref{eq:expected_activation}). Alternatively, the $\ell_2$-based importance score, $I_{\ell_2}$, incorporates the Root Mean Square over the dataset batches $\mathcal{B} \in \mathcal{D}$:\vspace{-4pt}
\begin{equation}
I_{\ell_2-FPS}(w_{k,j}^{(i)}) = |w_{k,j}^{(i)}| \cdot \sqrt{\frac{1}{|\mathcal{D}|} \sum_{\mathcal{B} \in \mathcal{D}} \sum_{x \in \mathcal{B}} (a_{k}^{(i-1)}(x))^2}.
\label{eq:ls-fps}
\end{equation}

\vspace{-4pt}As shown in \cref{tab:ablation_vtab}, the neuron-level scheme and $\ell_1$-based importance score consistently achieve leading performance across settings \cref{tab:ablation_vtab}a and \cref{tab:ablation_vtab}b. Consequently, we adopt the combination of neuron-level selection and the $\ell_1$ norm as our base FPS formulation.

\subsubsection{Impact of weight magnitude.}
To justify the inclusion of the weight magnitude in our formulation, we perform an ablation study where parameter importance is calculated based solely on the expected input activation. This corresponds to evaluating the $\ell_1$- and $\ell_2$-based importance scores defined in \cref{eq:fps_score} and \cref{eq:ls-fps}, respectively, without multiplying by the weight magnitude $|w_{k,j}^{(i)}|$.

The results in \cref{tab:ablation_vtab}c demonstrate a clear performance drop compared to the joint weight–activation approach in setting \cref{tab:ablation_vtab}b, empirically justifying the inclusion of the weight magnitude in our importance score calculation. Concluding these base design ablations, we formally adopt the combination of the $\ell_1$-norm and neuron-level selection as our standard FPS formulation.\vspace{-8pt}

\subsubsection{Impact of the GD mechanism.}
Building on the optimal base FPS configuration, we evaluate the GD mechanism. By leveraging the pre-trained model's behavior as a baseline, the GD mechanism prevents parameters from being selected if they consistently exhibit high-magnitude activations irrespective of the target task. To accurately quantify the activation growth, we ablate our proposed relative-growth formulation (see \cref{eq:gd_fps}) against an absolute-growth alternative. Rather than normalizing the magnitude change, the absolute-growth importance score simply computes the raw difference between the downstream and pre-trained expectations:\vspace{-2pt}
\begin{equation}
\label{eq:gd_fps_absolute}
I_{abs-GD}(w_{k,j}^{(i)}) = |w_{k,j}^{(i)}| \cdot \left( \mathbb{E}_{x \sim \mathcal{D}_{\text{down}}} \left[ |a_k^{(i-1)}(x)| \right] - \mathbb{E}_{x \sim \mathcal{D}_{\text{pre}}} \left[ |a_k^{(i-1)}(x)| \right] \right).
\end{equation}

\vspace{-2pt}As demonstrated in \cref{tab:ablation_vtab}d, our proposed relative-growth formulation not only surpasses the absolute-growth approach but also yields clear performance gains over the base FPS method, culminating in our final GD-FPS framework. This confirms that baseline normalization is essential for isolating task-sensitive parameters. Furthermore, we found that GD-FPS successfully limits cross-dataset parameter intersection—specifically reducing the average overlap of the base FPS on VTAB-1k by $5$\%—which indirectly validates our core design motivation.\vspace{-13pt}

\section{Conclusion}\vspace{-7pt}
\label{sec:conclusion}

In this paper, we propose Growth-Driven Feedforward Parameter Selection (GD-FPS), a novel gradient-free method that resolves the fundamental bottlenecks of Gradient-based Parameter Selection (GPS). By identifying critical parameters through task-specific activation growth relative to a pre-training anchor, GD-FPS bypasses the prohibitive costs of backpropagation. Our comprehensive evaluation across $26$ visual tasks demonstrates that GD-FPS achieves performance comparable to or even superior to the state-of-the-art PEFT baseline. Crucially, during the selection stage, our method reduces peak GPU memory usage by $18\times$ and accelerates execution by more than $2.7\times$ compared to GPS. Furthermore, by relying strictly on feedforward activations, GD-FPS ensures a deterministic selection process that is invariant to the stochasticity of batch sampling. This design demonstrates that high performance, efficiency, and structural robustness can coexist within selection-based PEFT, providing a scalable and practical solution for adapting large-scale pre-trained models in resource-constrained environments.



%
%
\bibliographystyle{splncs04}
\bibliography{main}

\clearpage
\appendix
\numberwithin{table}{section}
\numberwithin{figure}{section}

\end{document}